\RequirePackage[2020-02-02]{latexrelease}
\documentclass[twocolumn, switch]{article} 

\usepackage{preprint}

\usepackage{amsmath, amsthm, amssymb, amsfonts}

\usepackage[numbers,square]{natbib}
\usepackage[utf8]{inputenc}	
\usepackage[T1]{fontenc}	
\usepackage{xcolor}		
\usepackage[colorlinks = true,
            linkcolor = purple,
            urlcolor  = blue,
            citecolor = cyan,
            anchorcolor = black]{hyperref}	
\usepackage{booktabs} 		
\usepackage{nicefrac}		
\usepackage{microtype}		
\usepackage{lineno}		
\usepackage{float}			

\usepackage{lipsum}		
\usepackage{listings}   
\usepackage{pifont}     
\usepackage{algorithm2e} 

\definecolor{codegreen}{rgb}{0,0.6,0}
\definecolor{codegray}{rgb}{0.5,0.5,0.5}
\definecolor{codepurple}{rgb}{0.58,0,0.82}
\definecolor{backcolour}{rgb}{0.95,0.95,0.92}

\lstdefinestyle{codestyle}{
    backgroundcolor=\color{backcolour},   
    commentstyle=\color{codegreen},
    keywordstyle=\color{blue},
    numberstyle=\tiny\color{codegray},
    stringstyle=\color{codepurple},
    basicstyle=\ttfamily\footnotesize,
    breakatwhitespace=false,                         
    captionpos=b,                    
    keepspaces=true,                 
    numbers=left,                    
}

\usepackage{newfloat}
\DeclareFloatingEnvironment[name={Supplementary Figure}]{suppfigure}
\usepackage{sidecap}
\sidecaptionvpos{figure}{c}

\usepackage{titlesec}
\titlespacing\section{0pt}{12pt plus 3pt minus 3pt}{1pt plus 1pt minus 1pt}
\titlespacing\subsection{0pt}{10pt plus 3pt minus 3pt}{1pt plus 1pt minus 1pt}
\titlespacing\subsubsection{0pt}{8pt plus 3pt minus 3pt}{1pt plus 1pt minus 1pt}

\usepackage{tikz,xcolor,hyperref}

\definecolor{lime}{HTML}{A6CE39}
\DeclareRobustCommand{\orcidicon}{
	\begin{tikzpicture}
	\draw[lime, fill=lime] (0,0) 
	circle [radius=0.16] 
	node[white] {{\fontfamily{qag}\selectfont \tiny ID}};
	\draw[white, fill=white] (-0.0625,0.095) 
	circle [radius=0.007];
	\end{tikzpicture}
	\hspace{-2mm}
}
\foreach \x in {A, ..., Z}{\expandafter\xdef\csname orcid\x\endcsname{\noexpand\href{https://orcid.org/\csname orcidauthor\x\endcsname}
			{\noexpand\orcidicon}}
}

\title{Serverless Federated Learning with flwr-serverless}

\usepackage{xwatermark}

\usepackage{authblk}

\author[1]{Sanjeev V. Namjoshi}
\author[2]{Reese Green}
\author[3]{Krishi Sharma}
\author[4]{Zhangzhang Si}

\affil[1,2,3,4]{KUNGFU.AI}

\begin{document}

\twocolumn[ 
  \begin{@twocolumnfalse} 
  
\maketitle

\begin{abstract}
Federated learning is becoming increasingly relevant and popular as we witness a surge in data collection and storage of personally identifiable information. Alongside these developments there have been many proposals from governments around the world to provide more protections for individuals' data and a heightened interest in data privacy measures. As deep learning continues to become more relevant in new and existing domains, it is vital to develop strategies like federated learning that can effectively train data from different sources, such as edge devices, without compromising security and privacy. Recently, the Flower (\texttt{Flwr}) Python package was introduced to provide a scalable, flexible, and easy-to-use framework for implementing federated learning. However, to date, Flower is only able to run synchronous federated learning which can be costly and time-consuming to run because the process is bottlenecked by client-side training jobs that are slow or fragile. Here, we introduce \texttt{flwr-serverless}, a wrapper around the Flower package that extends its functionality to allow for both synchronous and asynchronous federated learning with minimal modification to Flower's design paradigm. Furthermore, our approach to federated learning allows the process to run without a central server, which increases the domains of application and accessibility of its use. This paper presents the design details and usage of this approach through a series of experiments that were conducted using public datasets. Overall, we believe that our approach decreases the time and cost to run federated training and provides an easier way to implement and experiment with federated learning systems.
\end{abstract}
\vspace{0.35cm}

  \end{@twocolumnfalse} 
] 



\section{Introduction}

Over the last few years, the decreased cost of data storage and increased usage of apps and digital technologies have led to an unprecedented surge in data collection and availability. This data collection revolution has proceeded alongside numerous advances in deep learning which has provided a usage for this data for a large variety of applications. The output predictions from these statistical models has become increasingly sophisticated and continues to have a direct impact on both society and the global economy. Notably, much of this data contains personal information, often collected directly from individuals and inferred from their behavioral patterns, or directly recorded in the form of digital healthcare data.

The widespread usage of this data in deep learning technologies has induced a greater interest and concern for data protection and privacy and a call for the codification of such protections under a legal framework. Recently, the European Union has enacted the General Data Protection Regulation (GDPR) \cite{gdpr, gpdr_url} which specifies the legality of personal data collection and usage as well as establishes the control of personal data as a human right. In the healthcare domain, collected medical data that contains sensitive information, electronic health records, is protected under the Health Insurance Portability and Accountability Act (HIPAA) in the United States which restricts and prohibits disclosure of this information to third parties without patient consent. 

Deep learning model training tasks generally require the aggregation of disparate data sources into one centralized location so it is fully accessible by the model. For example, one may wish to train a model using edge-devices belonging to individual users, which would necessitate combining this data together in one location. In the healthcare domain, medical images and patient clinical data, which may exist in different data centers, may need to be pooled to successfully train models representative of the general population. 

In response to the recent concerns around data privacy, and the legal requirements around protecting the interests of patients and end-users, researchers at Google introduced the federated learning framework \cite{fed_1, fed_2}. Federated learning addressed many of the data privacy and security concerns by allowing multiple datasets to be trained while located in separate locations so that they cannot be aggregated together. The local machines that contain both the data and the model are known as \textit{clients} and they connect to a central \textit{server} to aggregate weights to be used in the next round of training. Such decentralized training aims to solve the problems of data privacy and has been successfully applied to a number of different domains including applications in edge computing, such as Internet of Things networks \cite{fed_iot}, wireless computing \cite{fed_wireless}, and the healthcare domain \cite{fed_healthcare}. 

The open-source Flower Python package \cite{flower} was recently introduced to provide federated learning capabilities to a variety of different modeling frameworks and enable running on edge devices. Flower solves a number of different challenges to training federated models and running them in a production environment. Recently, Flower has been popular choice as a federated learning framework due to the simplicity of its lightweight design and flexibility. However, one challenge that has not yet been addressed is allowing Flower to run training asynchronously. At present, the Flower framework requires all connected clients to send their weights to the server before aggregation. Consequently, the next federated training round is delayed until all models have successfully completed their local epochs. When a client crashes due to out-of-memory or other common errors, the training needs to be restarted. Another operational pain point is managing the federation server. Often a separate server needs to be started for each training experiment, making it hard to scale to hundreds of experiments.

\begin{figure*}
    \centering
    \includegraphics{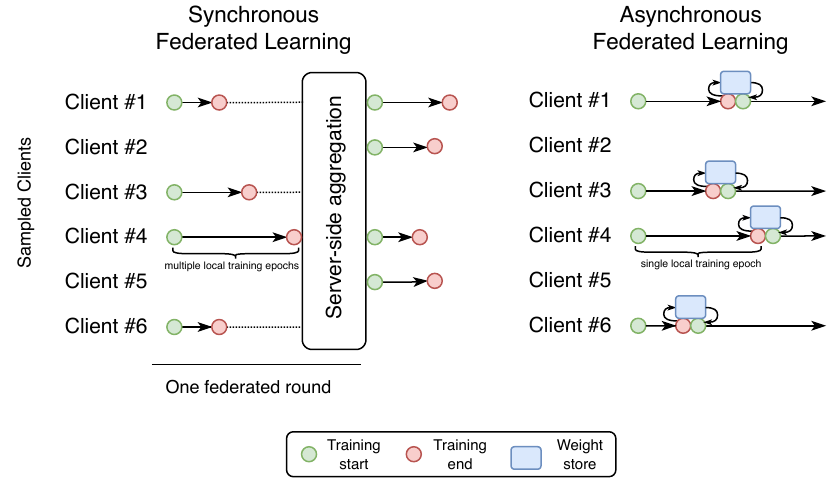}
    \caption{Synchronous versus asynchronous federated learning. In synchronous federated learning (left panel), a sampled set of clients begin multiple local training rounds. Upon completion, the client waits until the other clients finish. When all clients have finished, the server aggregates the weights and training continues. In our approach to asynchronous federated learning (right panel), the clients begin a single local training epoch and check a remote weight store for any weights deposited by any client that finished previously. It then downloads these weights locally, aggregates them, and continues training. Figure adapted from \cite{fed_async1}.}
    \label{fig:syncasync}
\end{figure*}

To address these issues, we introduce \texttt{flwr-serverless}, a wrapper around the Flower (flwr) framework that extends its capabilities to allow for both synchronous and asynchronous forms of federated learning without altering its core use pattern. Furthermore, the changes we make effectively allow Flower to be run "serverless" in the sense that weight aggregation occurs on the client side rather than the on the central server. The weights are updated from any accessible remote weight storage directory. Thus, our approach inherits all of the functionality and convenience of Flower while reducing the time and cost complexity related to updating the weights for the global model on the central server. Herein, we describe the architectural changes made to the federated learning workflow and the Flower package along with the asynchronous learning strategy that we employ. Finally, we demonstrate the results on some federated benchmark datasets. Our results show that asynchronous federated learning is robust, and in specific situations can significantly speed up federated training without sacrificing model performance.  

\section{Federated Learning}

Federated learning is a decentralized learning strategy that allows separate, private datasets stored across multiple devices or machines to be used to train a global model without requiring the model to access the dataset, or moving the dataset off the device into a central storage location. Each device that participates in federated learning is known as a \textit{client} which undergoes a number of \textit{local rounds} of training on its local dataset. After the client finishes its local rounds of training, it sends its current weights to a centralized \textit{server}, which stores the weights. All connected clients send their weights to this central server where they are averaged together by some federated aggregation \textit{strategy}. After aggregation, each client receives the newly aggregated weights and resumes training local rounds to update these weights.

Federated learning produces a unique scenario in that there are usually a large number of clients training at one time, but the data is likely to not be independent and identically distributed (\textit{i.i.d.}). Consequently, there are many types of federated aggregation strategies available aimed at dealing with these issues. In the base federated averaging strategy, known as FedAvg \cite{fed_2}, a random fraction of clients $C$ is chosen out of $K$ total clients where the client is indexed by $k$. For each global federated learning round $t=1, 2, \dots$, each randomly sampled client $k$ runs their model in parallel on their local datasets updating their local weights. The shared model weights $w_t$ are then updated at the next time step according to

\begin{equation}
    w_{t+1} \leftarrow \sum_{k=1}^K \frac{n_k}{n} w^k_{t+1}, 
\end{equation}

where $n$ denotes the index of the data point for client $k$. Thus, according to \cite{fed_2}, the FedAvg algorithm utilizes the general federated objective function,

\begin{equation}
    \min_{w \in \mathbb{R}^d},  \hspace{5mm} \text{where} \hspace{5mm} f(w) := \frac{1}{n} \sum_{i=1}^n f_i(w),
\end{equation}

where, $f_i(w)$ could be the typical supervised learning objective function $\ell(x_i, y_i; w)$ for each input/output sample pair in the dataset, indexed by $i$ with parameters $w$. In summary, FedAvg represents the average over all loss functions of the model parameters of each client. Many other federated strategies have been recently introduced and all are currently implemented in the Flower package \cite{flower}.

\paragraph{Synchronous versus asynchronous federated learning}

In most cases, federated learning is performed in synchronous fashion (\autoref{fig:syncasync}, left panel). Each client submits its weights to the server after completing a set number of local training rounds on its private data. The synchronization of all of the client's weights occurs once the last client to finish local training submits its weights to the server. Thus, synchronous federated learning implies that the overall training process is bottlenecked by the slowest client (\textit{stragglers}).

In order to address these concerns, an alternative approach is \textit{asynchronous} federated training (\autoref{fig:syncasync}, right panel).  A number of asynchronous federated learning strategies have been developed and implemented. The original FedAsync \cite{fedasync} strategy utilizes a mixing hyperparameter to control a client's contribution to the global aggregation based on its "staleness" (how slow the client is to complete its local training rounds). Other approaches have also used a similar staleness model including ASO-Fed \cite{asofed} and FedSa \cite{fedsa}. FedBuff \cite{fedbuff, fedbuff2} utilizes a buffered asynchronous aggregation approach which attempts to improve on secure aggregation protocols. The server selects a fraction of clients and aggregates them securely before updating. SAFA \cite{safa} uses a different approach in which a threshold of finished clients must be met before aggregation proceeds. PORT \citep{fed_async1} introduces another asynchronous update method which tries to balance between the staleness and communication cost by forcing stale clients to report their weights for aggregation after a threshold is met. Finally, a semi-synchronous federated learning paradigm has been explored in \cite{semi}.

\section{Serverless federated learning with \texttt{flwr-serverless}}

The \texttt{Flwr} package \cite{flower} has recently been developed to allow support for running federated learning on edge devices. In the Flower paper, the authors describe its design goals, which include being highly scalable, client-agnostic, communication-agnostic, privacy agnostic and flexible. Flower provides an easy-to-use package in which many standard machine learning frameworks (\texttt{TensorFlow}, \texttt{PyTorch}, \texttt{Scikit-learn} etc.) can be made into a federated client by wrapping a \texttt{fit} function. Despite these advantages, to date Flower does not offer support for asynchronous federated learning. Furthermore, we encountered a number of difficulties managing the communication between the main server doing the model weight aggregation and the clients with data sources stored in several cloud accounts with healthcare-related legal restrictions.

Here we introduce and provide an overview of \texttt{flwr-serverless}, our solution to expanding Flower's capabilities to include asynchronous federated learning, and allow client-side aggregation without a central server. We sought to develop the package with the following design principles in mind:

\begin{itemize}
    \item \textbf{Minimal modification}: We aim to retain the basic design goals and principles of the Flower package with minimal modification to its functionality or existing code and without interfering with its core design.
    \item \textbf{Serverless implementation}: Due to the numerous difficulties we encountered with launching and maintaining federated learning servers, we aim to provide asynchronous federated learning that can run in a \textit{serverless} fashion.
    \item \textbf{Flexibility}: Much like the Flower package, we aim for asynchronous federated learning to be compatible with machine learning frameworks such that it can be activated through callback functionality.
\end{itemize}

\paragraph{Design}

\begin{figure}
    \centering
    \includegraphics{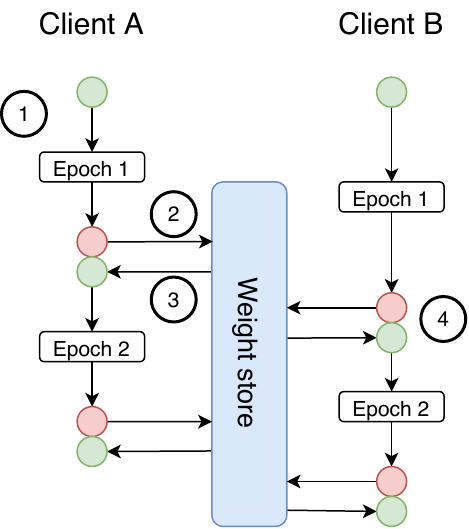}
    \caption{A detailed view of the asynchronous federated learning design in \texttt{flwr-serverless} showing two clients interacting with the weight store. In \ding{172}, Client A begins its first epoch. Upon completion, in \ding{173} it transfers its weights to the weight store. In \ding{174} the client downloads any available weights and aggregates them locally with its own weights according to a federated aggregation strategy. Client B follows a similar structure but in \ding{175}, we see that it takes longer to train one epoch. Therefore, a different set of weights may be available for aggregation than Client A depending on the other clients that have deposited their weights in the weight store. With many clients connecting, the weight store will contain a "running average" of the global weights proportional to the fastest clients that have finished epochs.}
    \label{fig:async-detail}
\end{figure}

In a typical synchronous federated learning experiment, each client runs for a number of local training training epochs before it sends its weights to a central server for aggregation. All participating clients are expected to submit their weights before the server aggregates the weights and broadcasts the new weights back to the clients. Our asynchronous implementation follows a different sequence of events (\autoref{fig:async-detail}). First, each client only completes a single local training epoch. Then, the client sends its weights to a remote \textit{weight store} and checks with the server to see if another client has recently deposited weights to this shared folder. If so, it downloads these weights and then aggregates them on the \textit{client side} and continuous training. The effect of this process is that the client effectively becomes serverless in the sense that the aggregation is performed by the client and not externally. In this system, the weight store is intended to be any remote folder that is accessible by the client machine, for example a bucket/blob location on a cloud service provider.

An interesting side effect of this kind of implementation is that each client may implement its own aggregation strategy. This opens up a number of new federating training possibilities and allows for further customization, especially for the stragglers who may average weights less frequently due to fewer connections to the shared folder. Furthermore this setup also implies that there is no "federated round". There are only local training rounds on each client with continuous weight updates between epochs if available. 

To run `flwr-serverless`, the user is expected to specify the following for each client:

\begin{itemize}
    \item The intended federated aggregation strategy to be used by the client.
    \item The location of the shared folder, for example, an AWS S3 bucket URI.
    \item The federated learning \textit{node} which has a specific strategy and shared folder.
    \item The \texttt{FlwrFederatedCallback} which will be passed to the framework's callback.
    \item The model \texttt{compile} and \texttt{fit} functions.
\end{itemize}

For example, with TensorFlow, the following is sufficient to launch and experiment with a single client. 

\begin{lstlisting}[language=Python]
# Create a FL Node that has a strategy and a shared folder.
from flwr.server.strategy import FedAvg
from flwr_serverless import AsyncFederatedNode, S3Folder

strategy = FedAvg()
shared_folder = S3Folder(directory="mybucket/experiment1")
node = AsyncFederatedNode(strategy=strategy, shared_folder=shared_folder)

# Create a keras Callback with the FL node.
from flwr.keras import FlwrFederatedCallback
num_examples_per_epoch = steps_per_epoch * batch_size # number of examples used in each epoch
callback = FlwrFederatedCallback(
    node,
    num_examples_per_epoch=num_examples_per_epoch,
)

# Join the federated learning, by fitting the model with the federated callback.
model = keras.Model(...)
model.compile(...)
model.fit(dataset, callbacks=[callback])
\end{lstlisting}

As more clients connect, they will all automatically update their weights, asynchronously, based on the presence of weights in the weight store.

\paragraph{Asynchronous update algorithm}

As a first proof of concept, we present the pseudocode for an asynchronous version of the original \texttt{FedAvg} algorithm called \texttt{FedAvgAsyc} (\autoref{alg:afedavg}). The notation for this algorithm follows the same notation from \cite{fed_2} which describes the original \texttt{FedAvg} algorithm in detail. Note that in this implementation, there is no server so all computations occur on the client side. First, all clients begin running in parallel with weights initialized at $w_0$. Then each client in $K$, indexed by $k$, runs a training epoch $i$ on the client side up to $E$ epochs. During this epoch, sampling occurs to see if this client will learn during this epoch. The probability of being sampled is controlled by the parameter $C$ (see below for more details on the meaning of sampling in this context). If a client is sampled, then it performs the \texttt{ClientUpdate} which entails computing the weight updates $w^k_i$ for client $k$ at epoch $i$ across all mini-batches $\mathcal{B}$, where $\mathcal{P}_k$ denotes the data point indexes of each client $k$. This weight is passed to the \texttt{WeightUpdate} procedure which pushes the weights to the weight store. The \textit{push} mechanism here pushes the weights using Flower for communication to the remote weight store. The client then performs a check to see if the remote server has changed state (as reported by a unique hash). The latest weights currently deposited by other asynchronous nodes would be contained here. These weights are then \textit{pulled} from the weight store to the local client in the array $\omega$. Client $k$ adds its weights $w^k$ to $\omega$ and then the weights are averaged according to $w_{i+1} \gets \sum_{k=1}^K \frac{n_k}{n} \omega[k]$ which become the new weight initialization for the next epoch by client $k$. If the client pulls weights from the weight store and finds that no weights are available, it resumes training on its current weights.

\RestyleAlgo{ruled}
\begin{algorithm}
    \caption{\texttt{FedAvgAsync}}\label{alg:afedavg}
    \SetKwBlock{DoParallel}{in parallel do}{end}
    \SetKwProg{Func}{Function}{:}{}
    \SetKwFunction{ClientUpdate}{ClientUpdate}
    \SetKwFunction{WeightUpdate}{WeightUpdate}
    $\text{initialize } w_0$\;
    \DoParallel{
        \ForEach{client $k$}{
            \ForEach{epoch $i $ from $1 $ to $ E$}{
                \If{$random[0,1] < C$}{
                    $w^k_i \gets \text{ClientUpdate}(k, w_i)$
                    $w^k_{i+1} \gets \text{WeightUpdate}(w^k_i)$
                }
            }
        }
    }
    
    \Func{\ClientUpdate{$k$, $w_i$}}{
        $\mathcal{B} \gets (\text{split} \hspace{1mm} \mathcal{P}_k \hspace{1mm} \text{into batches of size} \hspace{1mm} B)$\;
        \ForEach{batch $b \in \mathcal{B} $}{
            $w \gets w - \eta \nabla \ell(w; b)$
        }
        \Return $w$
    }
    \Func{\WeightUpdate{$w^k$}}{
        $\text{Push} \hspace{1mm} w^k \hspace{1mm} \text{to weight store}$\;
        $\text{Pull} \hspace{1mm} \omega \hspace{1mm} \text{from weight store}$\;
        $\omega[k] \gets w^k$\;
        $w_{i+1} \gets \sum_{k=1}^K \frac{n_k}{n} \omega[k]$\;
        \Return $w_{i+1}$
    }
\end{algorithm}

\paragraph{Differences between synchronous and asynchronous \texttt{FedAvg}}

There are a few key differences between the original $\texttt{FedAvg}$ algorithm and $\texttt{FedAvgAsync}$. First, since all computations take place asynchronously and simultaneously in parallel there is no need for a global federated round $t = 1,2, \dots$. Second, since clients either upload their weights after an epoch or they do not, there is no notion of a "local update"; all updates are local in this algorithm. Third, the notion of "sampling" must be handled differently because there is no global round over which sampling applies. In this case, sampling may be handled in one of two ways. Non-sampled clients can either wait for a set amount of time before resuming training or they can continue training without ever completing the \texttt{WeightUpdate} step. Fourth, due to the serverless nature of this algorithm, the \texttt{WeightUpdate} step may use a different type of update rule if preferred, opening the doors for other federated aggregation strategies to be utilized. Note that these algorithms could potentially be different for each client.

\paragraph{Synchronous serverless federated learning}

Note that we also provide the functionality to use synchronous federated learning in a serverless fashion. In this case, when clients are attempting to get parameters from other connected nodes, they must wait until all other clients have deposited their weights in the weight store. Then, all clients simultaneously download the weights $\omega$ and aggregate them on the client side.

\section{Experiments}

We designed a series of experiments to study the effects of several design choices on the quality of the model and training time. The design choices include:

\begin{itemize}
    \item the type of strategy: synchronous or asynchronous,
    \item the federated learning strategy, e.g FedAvg
    \item the number of federated nodes, and
    \item the label skew of the federated datasets on different nodes.
\end{itemize}

The experiments are performed in several datasets and machine learning tasks: MNIST \cite{mnist} digital classification, CIFAR 10 \cite{cifar} image classification, and language modeling on WikiText \cite{wikitext}.

\subsection{Data partitioning to simulate skew}

For all experiments, we split the dataset into a training set and a test set. The training set is partitioned according to the number of federated nodes. After training, the federated model is evaluated on the hold out test set without partitioning.

To simulate data disparity of federated datasets that may appear in practice, we simulate label skew for MNIST and CIFAR 10 datasets using the sampling procedure below. 

\begin{enumerate}
    \item The training examples are first partitioned into $n$ mutually exclusive subsets based on the label, where $n$ is the number of federated nodes. For example, when $n=2$, in MNIST digits 0-4 belong to the first partition and digits 5-9 belong to the second partition.

    \item Then, to simulate a skew of $s$ ($0 < s< 1$), with probability $s$ each training example is assigned to a node based on the partition; with probability $1-s$, the training example is assigned to a random node.
\end{enumerate}

\subsection{MNIST}

We run two experiments on the MNIST dataset. The first compares synchronous vs asynchronous learning on different data distributions. The second experiment compares different federated learning strategies on different numbers of nodes (2, 3, and 5).  The model architecture for all MNIST experiments is kept the same. It consists of two convolutional layers with max pooling and ReLU activation.
We used the Adam optimizer with a fixed learning rate of $1e^{-3}$, a batch size of 32, 1200 steps per epoch for 3 epochs. Model federation happened at the end of each epoch.

\subsubsection{Effect of synchronous vs asynchronous}

Below are our findings about the effect of synchronous vs. asynchronous strategies:

\begin{itemize}
    \item Asynchronous federation gets about the same accuracy as the synchronous counter-part, except when the skew of federated datasets become very large.
    \item Asynchronous federation is slightly faster than synchronous federation due to less waiting.
\end{itemize}

The training time of asynchronous federated learning tends to be shorter than its synchronous counterpart. This depends on how uneven the training speeds are on different federation nodes. For synchronous federation, when each node finishes its epoch within roughly at same time, there is little efficiency loss. Otherwise, the training speed of asynchronous federated learning is bottlenecked by the slowest node.

Besides the quantitative comparisons, we note that asynchronous federation offers clear operational benefits and robustness. In asynchronous federation, when a node fails, the other nodes keep working. While in synchronous training, the other nodes are stuck. This can be important as real world model training jobs can be fragile, due to code errors, out of memory issues, server crashes and many other reasons.

\begin{table}[H]
\begin{center}
\begin{tabular}{l|lll}
\toprule
    & \multicolumn{3}{c}{\textbf{Skew}} \\
    \textbf{Strategy} & \textbf{0} & \textbf{0.9} & \textbf{1}  \\
    \midrule
    
    sync & .987 $\pm$ .001 & .983 $\pm$ .002 & .894 $\pm$ .02 \\
    async & .985 $\pm$ .001 & .976 $\pm$ .003 & .734 $\pm$ .114 \\
    
    \bottomrule
   \end{tabular}
   \vspace{5pt}
\caption{Accuracy: comparing synchronous and asynchronous FedAvg strategies. Mean and 95\% confidence intervals are reported for repeated trials. For centralized training using the same hyper-parameters, the accuracy is 0.987.}
\label{tab:mnist1}
\end{center}
\end{table}

\autoref{tab:mnist1} show the validation accuracy with different federated strategies and number of nodes. Each epoch includes 1200 training steps with a batch size of 32. Three different data skews were tested: random (no skew), partial skew and full skew (no label overlap). Random partition is a simple random split of the data across the 2 nodes. For the full skew partition, each node gets a subset of the labels, digits 0-4 for node 1 and digits 5-9 for node 2 in this case. The partial skew attempts to model a more realistic skewed data, where the training data for each node consists of a different mixture the same labels. For node 1, the majority of examples are digits 0-4, while the remaining are digits 5-9. Node 2 has the opposite mixture.

While asynchronous federation can yield lower accuracy and higher variance of accuracy, it has clear operational benefits of robustness and speed. With synchronous federation, if participating nodes have different hardware capacity or training speed, the faster nodes stay idle while waiting for slower nodes, wasting valuable compute resources. Moreover, in practice large training jobs can crash for many different reasons. In synchronous federation, failure in one node can halt the whole federated training. Asynchronous federation is much more forgiving for individual node failures.

\subsubsection{Effect of federated strategy and the number of nodes}

Despite its simplicity, FedAvg performed well compared to FedAvgM and FedAdam. The accuracy of FedAvgM was consistently close to that of FedAvg, and no statistically significant improvement is observed. FedAdam resulted in consistently lower accuracy.

The number of nodes had significant effect on accuracy. More nodes resulted in lower accuracy. See details in \autoref{tab:mnist-0.9} and \autoref{tab:mnist-0.99}.

\begin{table}[ht]
    \centering
    \begin{tabular}{l|lll}
    \toprule
    & \multicolumn{3}{c}{\textbf{Number of Nodes}} \\
    \textbf{Strategy} & \textbf{2} & \textbf{3} & \textbf{5}  \\
    \midrule
    
    FedAvg & .983 $\pm$ .002 & .983 $\pm$ .001 & .979 $\pm$ .001 \\
    FedAvgM & .983 $\pm$ .001 & .983 $\pm$ .001 & .979 $\pm$ .001 \\
    FedAdam & .976 $\pm$ .002 & .97 $\pm$ .007 & .962 $\pm$ .007 \\
    \midrule

    FedAvg (async) & .976 $\pm$ .003 & .979 $\pm$ .002 & .97 $\pm$ .007 \\
    FedAvgM (async) & .981 $\pm$ .002 & .979 $\pm$ .001 & .971 $\pm$ .003 \\
    FedAdam (async) & .97 $\pm$ .005 & .928 $\pm$ .058 & .95 $\pm$ .012 \\
    
    \bottomrule
    \end{tabular}
    \vspace{5pt}
    \caption{Accuracy: comparing federated strategies with different number of nodes. Label skew of 0.9 is used in this experiment.}
    \label{tab:mnist-0.9}
    \end{table}

\begin{table}[ht]
    \centering
    \begin{tabular}{l|lll}
    \toprule
    & \multicolumn{3}{c}{\textbf{Number of Nodes}} \\
    \textbf{Strategy} & \textbf{2} & \textbf{3} & \textbf{5}  \\
    \midrule
    
    FedAvg & .975 $\pm$ .003 & .965 $\pm$ .002 & .949 $\pm$ .002 \\
    FedAvgM & .976 $\pm$ .002 & .965 $\pm$ .002 & .947 $\pm$ .001 \\
    FedAdam & .967 $\pm$ .003 & .95 $\pm$ .005 & .926 $\pm$ .006 \\
    \midrule
    FedAvg (async) & .971 $\pm$ .003 & .948 $\pm$ .005 & .928 $\pm$ .003 \\
    FedAvgM (async) & .967 $\pm$ .005 & .953 $\pm$ .009 & .925 $\pm$ .013 \\
    FedAdam (async) & .956 $\pm$ .014 & .91 $\pm$ .021 & .903 $\pm$ .015 \\
    \bottomrule
    \end{tabular}
    \vspace{5pt}
    \caption{Accuracy: comparing federated strategies with different number of nodes. Label skew of 0.99 is used in this experiment.}
    \label{tab:mnist-0.99}
\end{table}


\subsection{CIFAR 10}

On the CIFAR 10 dataset using ResNet-18 \cite{resnet}, we observed somewhat different behaviors of the federated strategies than in MNIST experiments:

\begin{itemize}
    \item Asynchronous FedAvg performed the same or better compared to the synchronous counter-part.
    \item Similar to MNIST experiments, higher skew resulted in significantly lower accuracy.
    \item The number of nodes also had a significant negative impact on accuracy.
\end{itemize}

The FedAdam strategy worked poorly for CIFAR 10 and is not shown in the tables.

For all experiments on CIFAR 10, we used the Adam optimizer with a fixed learning rate of $5e^{-4}$, a batch size of 128, 1200 steps per epoch for 20 epochs. Model federation happened at the end of each epoch. Detailed results are shown in \autoref{tab:cifar10-1}, \autoref{tab:cifar10-0.9} and \autoref{tab:cifar10-0.99}.

\begin{table}[H]
\begin{center}
\begin{tabular}{l|lll}
 \toprule
    & \multicolumn{3}{c}{\textbf{Skew}} \\
    \textbf{Strategy} & \textbf{0} & \textbf{0.9} & \textbf{1}  \\
    \midrule
    
     sync & .804 $\pm$ .003 & .744 $\pm$ .01 & .477 $\pm$ .014 \\
    async & .802 $\pm$ .004 & .753 $\pm$ .018 & .505 $\pm$ .048 \\
    
    \bottomrule
   \end{tabular}
   \vspace{5pt}
\caption{Accuracy: comparing synchronous and asynchronous FedAvg strategies for different levels of skew. For centralized training, the accuracy is 0.803.}
\label{tab:cifar10-1}
\end{center}
\end{table}

\begin{table}[ht]
    \centering
    \begin{tabular}{l|lll}
    \toprule
        & \multicolumn{3}{c}{\textbf{Number of Nodes}} \\
        \textbf{Strategy} & \textbf{2} & \textbf{3} & \textbf{5}  \\
        \midrule
        
        FedAvg & .744 $\pm$ .01 & .717 $\pm$ .005 & .69 $\pm$ .002 \\
        FedAvgM & .749 $\pm$ .002 & .715 $\pm$ .01 & .689 $\pm$ .004 \\
        \midrule

        FedAvg (async) & .753 $\pm$ .018 & .728 $\pm$ .003 & .692 $\pm$ .003 \\
        FedAvgM (async) & .733 $\pm$ .012 & .733 $\pm$ .006 & .689 $\pm$ .004 \\
        
        \bottomrule
    \end{tabular}
    \vspace{5pt}
    \caption{Accuracy: comparing federated strategies with different number of nodes. Label skew of 0.9 is used in this experiment.}
    \label{tab:cifar10-0.9}
\end{table}

\begin{table}[ht]
    \centering
    \begin{tabular}{l|lll}
    \toprule
        & \multicolumn{3}{c}{\textbf{Number of Nodes}} \\
        \textbf{Strategy} & \textbf{2} & \textbf{3} & \textbf{5}  \\
        \midrule
        
        FedAvg & .552 $\pm$ .019 & .545 $\pm$ .021 & .43 $\pm$ .026 \\
        FedAvgM & .566 $\pm$ .014 & .458 $\pm$ .006 & .441 $\pm$ .022 \\
        \midrule

        FedAvg (async) & .615 $\pm$ .044 & .577 $\pm$ .024 & .418 $\pm$ .03 \\
        FedAvgM (async) & .651 $\pm$ .011 & .564 $\pm$ .012 & .433 $\pm$ .028 \\
        
        \bottomrule
    \end{tabular}
    \vspace{5pt}
    \caption{Accuracy: comparing federated strategies with different number of nodes. Label skew of 0.99 is used in this experiment.}
    \label{tab:cifar10-0.99}
\end{table}

\subsection{Language modeling on WikiText}

Training large language models often requires high end GPUs (such as A100, H100) with much higher memory than previous generations of GPUs. This makes it harder for researchers and practitioners to experiment with recipes for high quality language models.

Serverless federated learning can help reduce the barrier to entry for language modeling experiments. To illustrate the idea, we ran experiments on WikiText-103-V1 using a modest-sized open source LLM \textrm{Pythia-14M} \cite{pythia}.

In all the WikiText experiments, we used the first $100,000$ examples in the training set and report the next token prediction accuracy in the first $1,000$ examples in the validation set. We used the AdamW \cite{adamw} optimizer with a learning rate of $2e^{-5}$, a batch size of 16 with gradient accumulation every 10 steps. The model was trained for 3 epochs and federated aggregation happened at the end of each epoch.

\begin{table}[H]
\begin{center}
\begin{tabular}{l|lll}
\toprule
    & \multicolumn{3}{c}{\textbf{Number of Nodes}} \\
    \textbf{Strategy} & \textbf{2} & \textbf{3} & \textbf{5}  \\
    \midrule
    
    FedAvg & .26 $\pm$ .002 & .237 $\pm$ .004 & .227 $\pm$ .008 \\
    FedAvg (async) & .251 $\pm$ .005 & .239 $\pm$ .006 & .221 $\pm$ .006 \\
    \bottomrule
\end{tabular}
   \vspace{5pt}
\caption{Accuracy: comparing synchronous and asynchronous FedAvg strategies for different number of nodes. For centralized training, the accuracy is 0.279.}
\label{tab:wikitext}
\end{center}
\end{table}

\autoref{tab:wikitext} shows the accuracy for next token prediction is roughly the same between synchronous and asynchronous FedAvg strategies. Similar to previous experiments, the accuracy decreases with more federated nodes being used. 

This implies a trade-off between training speed and accuracy. Using more nodes is expected to be faster in processing training data, at the cost of lower accuracy. Better federation algorithms are expected to help reduce the gap.

\section{Limitations and future work}

In its current state, \textrm{flwr-serverless} represents a modest exploration of practical federated learning. Our current experiments cover a limited portion of design space. Much more is to be explored:

\begin{enumerate}
    \item We did not study the effect of model architectures. It is hard to tell if the different behaviors of the federated strategies were the result of a different dataset (task), or the result of a different model architecture.
    \item We did not implement staleness-aware asynchronous strategies (e.g. \cite{buffered-async}) that were shown to produce higher accuracy.
    \item We only experimented with a limited set of federation strategies from flwr: FedAvg, FedAvgM and FedAdam. 
    \item We did not study the effect of frequency to federation.
    \item We did not explore working with more than five nodes.
    \item We simulated concurrent training jobs with python multi-threading, which may have subtle differences from federated learning in fully isolated processes.
\end{enumerate}

Besides going beyond the above limitations, it is also interesting to explore partial model updates (e.g. \cite{partial-model}) in federated learning, which enables training much larger models than what fit into an individual node's GPU memory.

Despite the current limitations, we believe \texttt{flwr-serverless} can address many operational pain points for researchers and practitioners. We are excited about its potential to scale to large datasets and large models using a fleet of affordable compute nodes.

\section{Reproducibility}

The source code is available in Github\footnote{\url{https://github.com/kungfuai/flwr_serverless}}. The experiments and model weights are stored in Weights and Biases and will be shared in the same github repo.

\section{Conclusion}

We introduced \texttt{flwr-serverless}, an enhancement to the Flower package, to facilitate synchronous and asynchronous federated learning without a central server. Through our experimentation, it is evident that our approach provides a promising avenue for enhancing efficiency in federated learning setups. While our experiments were restricted in scope, \texttt{flwr-serverless} serves as a practical tool that simplifies the intricacies of federated learning, ensuring it remains accessible and practical even as the datasets and models continue to grow. We are optimistic about the future contributions and refinements in this direction.



\normalsize
\bibliography{references}


\end{document}